\pdfoutput=1
\documentclass[letterpaper,10pt,conference]{ieeeconf}

\IEEEoverridecommandlockouts
\usepackage{graphics}
\usepackage{epsfig}
\usepackage{amsmath}
\usepackage{amssymb}
\usepackage{amsfonts}
\usepackage{multirow}
\usepackage[table]{xcolor}
\usepackage{bm}
\usepackage{booktabs}
\usepackage{makecell}
\usepackage{nicefrac}
\usepackage{graphicx}
\usepackage{colortbl}
\usepackage{algorithm}
\usepackage{algorithmic}
\usepackage[labelfont=bf,labelsep=period]{caption}
\usepackage{subcaption}
\usepackage{comment}
\usepackage{thmtools,thm-restate}
\usepackage{wrapfig}

\usepackage{tabularx}
\usepackage{float}
\usepackage{dblfloatfix}
\usepackage{placeins}
\usepackage{url}
\usepackage{microtype}
\makeatletter
\AtBeginDocument{\renewcommand{\fnum@table}{Table~\thetable}}
\let\NAT@parse\undefined
\makeatother
\usepackage[breaklinks,colorlinks,citecolor=green]{hyperref}
\usepackage[capitalise]{cleveref}
\usepackage{cite}

\title{\LARGE \bf
READ: Learning Risk-Informed Fields for End-to-End Autonomous Driving
}

\author{Zhiyuan Liu$^{1*}$, Yuanxin Tian$^{1*}$, Zehong Ke$^{1}$, Jinhao Li$^{1}$,\\
Hao Cheng$^{1}$, Zhenhua Xu$^{1,\dagger}$, Wenhao Yu$^{1,\dagger}$, Jianqiang Wang$^{1}$%
\thanks{$^{1}$ School of Vehicle and Mobility, Tsinghua University.}%
\thanks{$^{*}$ Equal contribution.}%
\thanks{$^{\dagger}$ Corresponding authors: Zhenhua Xu
  ({\tt\footnotesize zxubg@connect.ust.hk}) and Wenhao Yu
  ({\tt\footnotesize eyre530056@gmail.com}).}%
}

\begin{document}

\maketitle
\thispagestyle{empty}
\pagestyle{empty}

\begin{abstract}
Autonomous driving requires more than recognizing what is present in a scene: a planner must determine how road structure, surrounding agents, and their motion states should influence a future maneuver. Existing learning-based planners can capture these influences through latent scene features and trajectory decoders, but the relationship between environmental factors and candidate actions often remains implicit. This limits the ability to inspect, diagnose, or refine how scene context affects the safety of a predicted trajectory. Classical safety fields provide an explicit spatial representation of this relationship, but their risk shapes and relative weights are prescribed in advance and do not adapt to each scene. We introduce READ, a framework that learns an explicit, planning-aligned risk representation from complementary geometric and behavioral constraints. READ instantiates this representation as a continuous spatiotemporal field, enabling differentiable queries along candidate trajectories. The learned field connects scene understanding with action selection by encouraging predicted trajectories to align with low-risk regions, while retaining a differentiable interface for trajectory evaluation and refinement. READ integrates with both end-to-end planners and Vision-Language-Action models. Experiments on NAVSIM show consistent gains across matched end-to-end backbones and strong performance in a VLA setting; READ also achieves competitive results on NAVSIM v2. These results establish learned spatial risk as an explicit, adaptable representation for safe planning.
\end{abstract}

\section{Introduction}
\label{sec:intro}

Autonomous driving requires systems to operate safely in complex, dynamic environments, where decisions depend on understanding which regions are safe to traverse and how that changes over time. Beyond recognizing what is present in a scene, a planner must determine how surrounding structure and motion should influence a future maneuver.

Classical approaches provide one way to make this relationship explicit by formulating safety as a cost over candidate behaviors. Obstacles, road structure, and interactions with other agents are translated into penalties distributed over space and time. Artificial potential fields model such a landscape through attractive and repulsive forces~\cite{Khatib1986}. Driving safety fields further unify driver--vehicle--road interactions and describe how multiple sources of risk propagate through a traffic scene~\cite{safetyfield}. Their continuous form is intuitive and useful for planning because a trajectory can be evaluated or optimized directly on the field. In practice, however, these formulations depend on carefully designed risk terms and globally selected parameters. The amplitude, spatial decay, anisotropy, and relative strength of different sources must be specified in advance, although suitable margins vary with road geometry and dynamic interactions.

Modern learning-based approaches, in contrast, learn representations and policies directly from data. End-to-end systems commonly construct bird's-eye-view features and occupancy predictions~\cite{UniAD,TransFuser}, world models predict how the scene may evolve~\cite{GAIA-1,FIERY,OccWorld}, and Vision-Language-Action models introduce semantic and latent reasoning~\cite{DriveVLM,DriveLM,FSDrive,Latent-WAM,li2026sgdrive}. These representations provide effective geometric, temporal, and semantic priors for planning, but the relationship between environmental factors and their influence on candidate actions often remains implicit in latent scene features and trajectory decoders. A complementary question is whether a learned intermediate representation can make this influence more explicit while remaining adaptive to scene context. Locations with the same semantic label can have different motion implications. A free region beside a fast-moving vehicle is not equivalent to one in an empty lane, and an occupied cell alone does not describe how far that agent should influence nearby trajectories. A behavior-aware spatial representation can expose this distinction and regularize shared features around planning-relevant structure.

Learning-based navigation has also revisited potential functions. One-4-All learns a geodesic potential over visual embeddings for graph-free image-goal navigation~\cite{NPF}, while FlowDrive introduces risk-potential and lane-attraction fields as structured BEV priors for driving~\cite{jiang2025flowdrive}. These works show that field structure can benefit learned planning. In driving, however, a field defined by analytic rules still carries manually selected amplitudes, spatial decay, anisotropy, and source weights across scenes. This motivates learning scene-dependent field geometry while retaining the continuity and planning utility of classical formulations.

\begin{figure*}[t]
\centering
\includegraphics[width=0.98\textwidth]{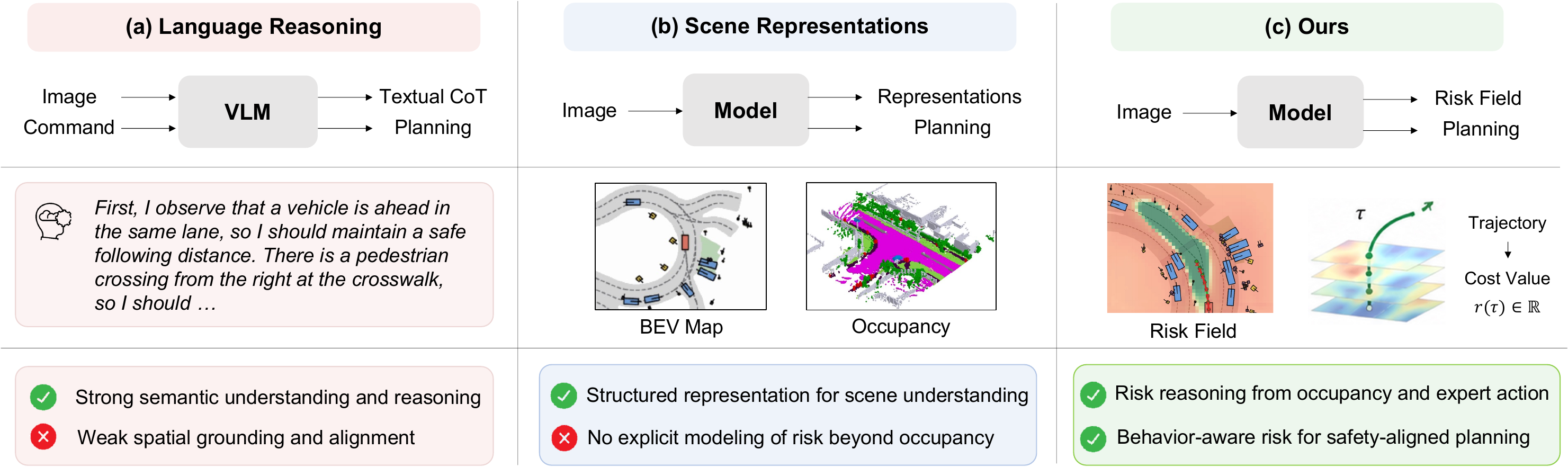}
\vspace{2mm}
\caption{Reasoning representations for autonomous driving. READ complements learned scene representations with a continuous, behavior-aware notion of spatial risk. Geometric and behavior-derived constraints allow the field to adapt its spatial influence to each scene.}
\label{fig:teaser}
%\vspace{-2mm}
\end{figure*}

We introduce READ, a framework that learns an explicit, planning-aligned risk representation from complementary geometric and behavioral constraints. READ instantiates this representation as a continuous spatiotemporal field, enabling differentiable queries along candidate trajectories. READ builds on classical field-based formulations, but replaces hand-crafted field geometry with a learnable representation that captures environmental structure and interaction dynamics from data. Non-drivable regions and future agent occupancy provide reliable geometric anchors, while expert trajectories and deviations from inertial motion contribute behavior-dependent evidence. Rather than reproducing a complete surface generated by fixed rules, the shared representation reconciles these constraints with trajectory supervision. The field can therefore adapt its spatial support to each scene while retaining the continuity and locality useful for planning, including different levels of influence in physically unoccupied regions.

As an explicit inductive bias, the representation regularizes the relationship between spatial structure and action selection, encouraging predicted trajectories to align with low-risk regions. READ is architecture-agnostic. In traditional end-to-end models, it augments shared spatial features with structured risk supervision. In Vision-Language-Action models, it forms risk-aware latent tokens that interact with action queries. Beyond representation learning, the differentiable field can be evaluated along trajectories and can optionally guide selective trajectory refinement without retraining the planner.

We summarize our contributions as follows:
\begin{itemize}
    \item \textbf{We introduce an explicit, planning-aligned risk representation for autonomous driving}, bridging classical field-based safety formulations with modern data-driven planning. READ instantiates this representation as a continuous, behavior-aware field beyond occupancy.
    \item \textbf{We show that READ transfers across planner types}, integrating the learned field into end-to-end and Vision-Language-Action models, and evaluating its use for trajectory assessment and optional differentiable refinement.
    \item \textbf{We evaluate READ on NAVSIM}, showing consistent gains with matched end-to-end backbones and strong results in a VLA setting.
\end{itemize}

\section{Related Work}
\label{sec:related}

\textbf{End-to-end autonomous driving.}
Early end-to-end driving systems learned direct mappings from sensory observations to control~\cite{ALVINN,CIL}. Recent work uses transformer-based sensor fusion~\cite{TransFuser}, jointly optimizes perception, prediction, and planning~\cite{UniAD}, or generates multimodal trajectories~\cite{DiffusionDrive,Hydra-MDP}. Predictive representations expose more of the future scene through video or occupancy modeling~\cite{GAIA-1,FIERY,OccWorld}. These approaches learn powerful planning features, but spatial risk is usually represented implicitly or only through geometric occupancy. READ is complementary: it adds a continuous cost representation that can both regularize a shared feature space and be queried along arbitrary trajectories.

\textbf{Vision-Language-Action models for driving.}
VLA models bring large-scale semantic pretraining and intermediate reasoning to autonomous driving~\cite{DriveVLM,DriveLM}. Existing approaches use textual chains, visual spatiotemporal reasoning, latent world-action models, or scene-to-goal hierarchies to connect scene understanding with planning~\cite{Reason2Drive,FSDrive,Latent-WAM,li2026sgdrive}. READ complements these mechanisms with risk-structured latent tokens that retain a continuous spatial meaning for planning.

\textbf{Safety fields and learned planning costs.}
APFs represent obstacles and goals through continuous repulsive and attractive potentials~\cite{Khatib1986,Borenstein1989}. Driving safety fields extend this view to road topology, vehicle motion, and interactions~\cite{APF1,safetyfield}. Related semi-stochastic fields learn representations of multi-vehicle interactions~\cite{APF1}. The central insight remains useful: safety is not only occupied versus free, but a spatial influence that varies with context and can be accumulated along a path. Classical formulations make this influence explicit, although their kernels and coefficients must be selected for the expected scene distribution.

One-4-All learns a geodesic potential over visual embeddings for image-goal navigation~\cite{NPF}, and FlowDrive uses risk-potential and lane-attraction fields as structured BEV priors for diffusion planning and anchor refinement~\cite{jiang2025flowdrive}. These methods establish field structure as a useful prior for learned planning. READ follows this direction while learning scene-conditioned field geometry from geometric and behavioral constraints rather than prescribing a complete analytic surface.

\section{Method}
\label{sec:method}

READ models the driving scene as a continuous spatiotemporal risk field and learns it jointly with the planning policy. The field is trained from complementary geometric and behavioral evidence, integrated with two representative planning architectures, and optionally used to refine trajectories.

\begin{figure*}[t]
	\centering
	\includegraphics[width=0.98\textwidth]{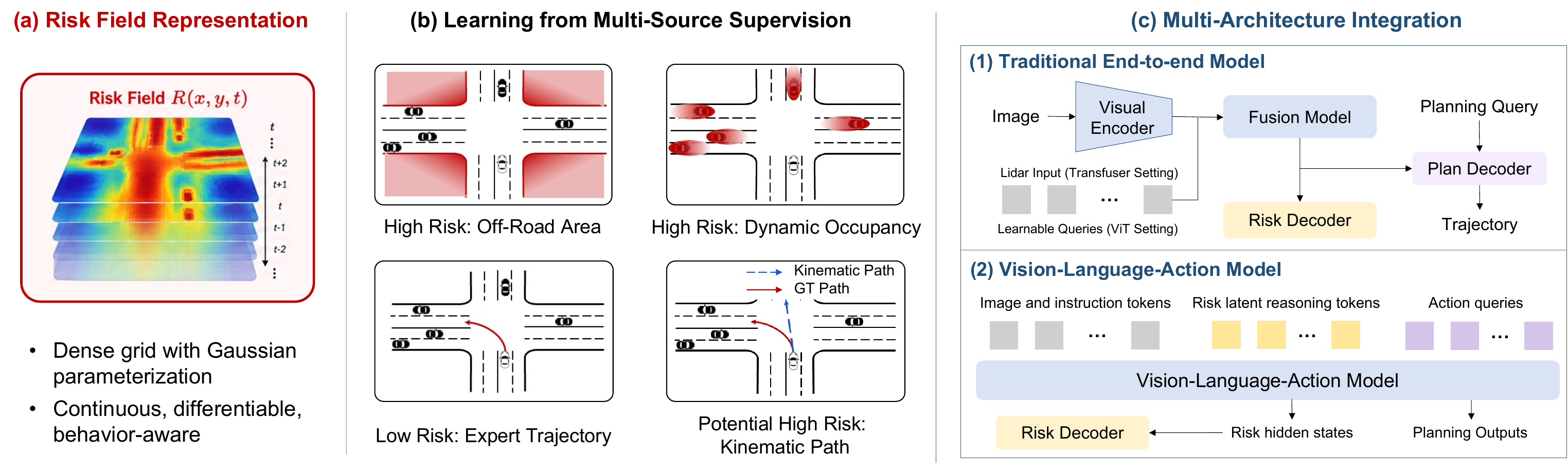}
	\caption{\textbf{READ overview.} Partial geometric and behavior-derived constraints supervise a continuous Gaussian risk field. The same representation supplies structured gradients to end-to-end and VLA planners and can optionally refine a predicted trajectory.}
	\label{fig:pipeline}
\end{figure*}

\subsection{Preliminaries}
\label{sec:prelim}

\noindent\textbf{Problem formulation.}
In the NAVSIM setting~\cite{dauner2024navsim}, the agent receives sensor observations, ego state, and a high-level navigation command $c$. The planner predicts future ego-centric BEV waypoints $\tau=\{(x_k,y_k)\}_{k=1}^{K}$ over horizon $T_{\mathrm{plan}}$. We seek an intermediate representation that improves this prediction while retaining a direct geometric meaning outside the planner's latent space.

\noindent\textbf{Field construction across paradigms.}
Classical safety fields convert objects, lanes, and boundaries into a continuous cost through analytic kernels~\cite{safetyfield}. Their smoothness supports both trajectory comparison and gradient-based optimization. Recent learning-based planners reuse this structure as a model prior. FlowDrive, for example, introduces risk-potential and lane-attraction fields for anchor refinement and diffusion planning~\cite{jiang2025flowdrive}. Such explicit priors connect spatial structure with planning, but their field geometry is still determined by the adopted analytic design.

READ retains continuity, differentiability, and spatial locality without supervising one complete hand-designed surface. Maps, future agent occupancy, demonstrated paths, and kinematic residuals constrain selected values, while the remaining geometry is learned jointly with planning. Thus the distinction is between fitting a complete analytic field and learning an adaptive field from partial constraints. Both use structured prior knowledge. READ places fewer assumptions on how risk must decay around each source.

\subsection{Adaptive Spatiotemporal Risk Field}
\label{sec:risk_field}

We model risk as a time-dependent field $\mathcal{R}(x,y,t)$ in BEV. The field is constrained where training data provide defensible safety evidence; it is not assigned a dense analytic target at every location.

\subsubsection{Gaussian Parameterization}
\label{sec:gaussian_param}

We place anisotropic Gaussian primitives on an $H\times W$ grid. For cell $(i,j)$ and future time $t$, the center $\boldsymbol{\mu}_{ij}=(x_{ij},y_{ij})$ is fixed, while weight $w_{ijt}\in[0,1]$ and diagonal scales are predicted:
\begin{equation}
\label{eq:gaussian}
\resizebox{0.88\columnwidth}{!}{$\displaystyle
\mathcal{G}_{ijt}(x,y)=\frac{1}{2\pi\sigma_{x,ijt}\sigma_{y,ijt}}\exp\!\left(-\frac{(x-x_{ij})^2}{2\sigma_{x,ijt}^2}-\frac{(y-y_{ij})^2}{2\sigma_{y,ijt}^2}\right).$}
\end{equation}
The continuous field is
\begin{equation}
\label{eq:risk_field}
\mathcal{R}(x,y,t)=\sum_{i=1}^{H}\sum_{j=1}^{W}w_{ijt}\mathcal{G}_{ijt}(x,y).
\end{equation}
This parameterization guarantees a smooth and differentiable query function. Learned scales let the same basis express tight boundaries near static structure or broader influence around uncertain moving agents, while time-indexed weights and scales allow the landscape to evolve with predicted interactions. We denote grid samples by $\mathcal{C}_{ij}^{t}=\mathcal{R}(x_{ij},y_{ij},t)$.

\subsubsection{Partial Safety Constraints}
\label{sec:supervision}

Unlike occupancy-only supervision, driving risk also reflects demonstrated preferences: a physically free region may still be avoided because of nearby motion or an emerging interaction. We therefore combine five losses. Map and agent annotations provide objective geometric anchors, demonstrated motion supplies behavior-derived constraints, and sparsity prevents a trivial high-cost field. Together they shape the field without assigning a target value to every location.

\paragraph{Static constraint.}
For non-drivable cells $\mathcal{S}$ rasterized from the map, we impose a high target $r_0$ on the time-averaged field $\bar{\mathcal{C}}_{ij}$:
\begin{equation}
\mathcal{L}_{\mathrm{ndr}}=\frac{1}{|\mathcal{S}|}\sum_{(i,j)\in\mathcal{S}}\|\bar{\mathcal{C}}_{ij}-r_0\|^2.
\label{eq:loss_ndr}
\end{equation}

\paragraph{Dynamic constraint.}
Let $\mathcal{D}_t$ contain cells covered by projected future agent boxes. We require high risk only at these annotated time-specific locations:
\begin{equation}
\mathcal{L}_{\mathrm{agent}}=\frac{1}{\sum_t|\mathcal{D}_t|}\sum_t\sum_{(i,j)\in\mathcal{D}_t}\|\mathcal{C}_{ij}^{t}-r_0\|^2.
\label{eq:loss_agent}
\end{equation}

\paragraph{Global sparsity.}
Positive constraints alone admit the degenerate solution of high risk everywhere. We use
\begin{equation}
\mathcal{L}_{\mathrm{reg}}=\frac{1}{HWT}\sum_{i,j,t}\mathcal{C}_{ij}^{t}
\label{eq:loss_reg}
\end{equation}
as an optimization prior that discourages unsupported activation. It should not be interpreted as ground-truth safety outside annotated regions.

\paragraph{Demonstrated-path constraint.}
Under the imitation-learning assumption that recorded expert motion is locally feasible, the field should be low at expert waypoints $\mathcal{P}$:
\begin{equation}
\mathcal{L}_{\mathrm{gt}}=\frac{1}{|\mathcal{P}|}\sum_{(x_k,y_k,t_k)\in\mathcal{P}}\|\mathcal{R}(x_k,y_k,t_k)\|^2.
\label{eq:loss_gt}
\end{equation}
This is a sampled behavioral constraint, not a claim that every unvisited location is unsafe.

\paragraph{Kinematic-residual constraint.}
A constant-velocity rollout $\mathbf{p}^{\mathrm{kin}}(t)=\mathbf{v}t$ supplies a counterfactual query path. Its distance from the demonstration, $r_t=\|\mathbf{p}^{\mathrm{kin}}_t-\mathbf{p}^{\mathrm{gt}}_t\|$, is normalized using the training-set 90th-percentile scale $s_t$ and matched at the queried point:
\begin{equation}
\mathcal{L}_{\mathrm{res}}=\frac{1}{T}\sum_t\|\mathcal{R}(\mathbf{p}^{\mathrm{kin}}_t,t)-\hat r_t\|^2,
\quad \hat r_t=\operatorname{clamp}(r_t/s_t,0,1).
\label{eq:loss_res}
\end{equation}
Because intentional turns can create a large residual without indicating a hazard, we down-weight this term according to heading change. The complete field loss is
\begin{equation}
\mathcal{L}_{\mathrm{field}}=\lambda_n\mathcal{L}_{\mathrm{ndr}}+\lambda_a\mathcal{L}_{\mathrm{agent}}+\lambda_r\mathcal{L}_{\mathrm{reg}}+\lambda_g\mathcal{L}_{\mathrm{gt}}+\lambda_k\mathcal{L}_{\mathrm{res}}.
\end{equation}
The coefficients balance the supervision signals. Spatial influence is determined by the predicted Gaussian weights and scales rather than fixed directly by these loss weights.

This separation is useful in mixed traffic. The static and dynamic masks identify locations where a high value is defensible, but they do not determine how far influence should extend into nearby free space. Expert and residual queries add evidence along two behaviorally meaningful paths rather than filling the entire map. Joint prediction must therefore reconcile all constraints in a common surface. The Gaussian basis supplies smooth interpolation, while learned weights and scales allow the support to change with object arrangement, time horizon, and the shared planning context.

\subsection{Integration with Driving Models}
\label{sec:integration}

We use the same field definition in two architectures to assess its utility beyond a single planner design.

\subsubsection{Risk-Field Auxiliary Learning for E2E Planning}
\label{sec:e2e_integration}

For BEV-based planners, a convolutional decoder predicts $\{w_{ijt},\sigma_{x,ijt},\sigma_{y,ijt}\}$ from the shared BEV state, while a trajectory query cross-attends to that state. Training uses
\begin{equation}
\mathcal{L}_{\mathrm{e2e}}=\mathcal{L}_{\mathrm{traj}}+\alpha\mathcal{L}_{\mathrm{field}}.
\end{equation}
The field is therefore an auxiliary structured objective: it supplies dense spatial gradients to the shared representation, while the trajectory decoder remains a direct prediction head at inference.

\subsubsection{Risk-Structured Latent Supervision for VLA}
\label{sec:vla_integration}

The VLA model augments visual and language tokens with risk latents $\mathbf{Z}_r$ and trajectory queries $\mathbf{Z}_a$. Risk tokens are processed jointly with the scene context and their output states $\mathbf{H}_r$ are decoded into $\mathcal{R}$. Trajectory queries then attend to both contextual tokens and $\mathbf{H}_r$ before waypoint and score prediction~\cite{guo2025ipaditerativeproposalcentricendtoend,li2025drivevla}. The objective is
\begin{equation}
\mathcal{L}_{\mathrm{VLA}}=\mathcal{L}_{\mathrm{traj}}+\beta\mathcal{L}_{\mathrm{field}}.
\end{equation}
The field loss propagates through the spatial decoder into $\mathbf{H}_r$, while trajectory supervision aligns the action queries with expert behavior. This realizes latent risk-aware reasoning without requiring textual chain-of-thought annotations and tests whether the field representation transfers beyond BEV-only architectures.

\subsection{Differentiable Trajectory Refinement}

The primary use of READ is joint representation learning. Its continuous form additionally permits post-hoc guidance for a trajectory produced by READ or an external planner. Given $\tau_0=\{(x_k^0,y_k^0,\theta_k^0)\}_{k=1}^{K}$, we refine only trajectories whose initial average risk exceeds threshold $\rho$ by minimizing
\begin{equation}
\label{eq:refine}
\mathcal{L}_{\mathrm{refine}}=\lambda_{\mathrm{risk}}\mathcal{L}_{\mathrm{risk}}+\lambda_{\mathrm{traj}}\mathcal{L}_{\mathrm{traj}},
\end{equation}
where
\begin{equation}
\mathcal{L}_{\mathrm{risk}}=\frac{1}{K}\sum_{k=1}^{K}\mathcal{R}(x_k,y_k,t_k),
\end{equation}
and $\mathcal{L}_{\mathrm{traj}}$ preserves the initial prediction while enforcing heading consistency, waypoint smoothness, acceleration and jerk regularity, and forward progress. Optimization is performed directly on waypoint positions and headings without re-querying or retraining the policy. This separates the representation-learning benefit from optional post-processing, and the two uses are evaluated independently.

\FloatBarrier
\begin{table*}[!t]
\centering
\captionsetup{width=0.95\textwidth}
\caption{Planning results on the original NAVSIM benchmark. The rightmost PDMS is the primary aggregate metric. The representation column denotes each method's explicit planning-oriented intermediate representation; READ uses risk fields rather than dense semantic BEV supervision. Gray rows add READ to the corresponding architecture. Drive-JEPA$^{\dagger}$ is our reproduced single-modal BEV planner.}
\label{tab:navsim_main}
\scriptsize
\renewcommand{\arraystretch}{1.12}
\setlength{\tabcolsep}{2.8pt}
\begin{tabularx}{0.95\textwidth}{@{}>{\raggedright\arraybackslash}p{2.45cm}>{\centering\arraybackslash}p{1.45cm}*{5}{>{\centering\arraybackslash}X}|>{\centering\arraybackslash}X@{}}
\toprule
Method & Rep. & NC$\uparrow$ & DAC$\uparrow$ & TTC$\uparrow$ & C$\uparrow$ & EP$\uparrow$ & \textbf{PDMS}$\uparrow$ \\
\midrule
\multicolumn{8}{l}{\emph{Traditional end-to-end methods}} \\
TransFuser~\cite{TransFuser} & BEV & 97.7 & 92.8 & 92.8 & 100.0 & 79.2 & 84.0 \\
\rowcolor{gray!15} TransFuser + READ & Risk field & 97.8 & 93.3 & 93.0 & 100.0 & 79.6 & 84.6 \\
Hydra-MDP~\cite{Hydra-MDP} & -- & 98.3 & 96.0 & 94.6 & 100.0 & 78.7 & 86.5 \\
DiffusionDrive~\cite{DiffusionDrive} & BEV & 98.2 & 96.2 & 94.7 & 100.0 & 82.2 & 88.1 \\
WoTE~\cite{Li_2025_ICCV} & BEV & 98.5 & 96.8 & 94.9 & 99.9 & 81.9 & 88.3 \\
Drive-JEPA$^{\dagger}$~\cite{wang2026drive} & BEV & 98.7 & 96.7 & 95.4 & 100.0 & 83.2 & 89.1 \\
\rowcolor{gray!15} Drive-JEPA$^{\dagger}$ + READ & Risk field & \textbf{99.0} & \textbf{97.2} & \textbf{96.4} & \textbf{100.0} & \textbf{83.8} & \textbf{90.0} \\
\midrule
\multicolumn{8}{l}{\emph{Vision-language-action methods}} \\
AutoVLA~\cite{zhou2025autovla} & Language & 98.4 & 95.6 & \textbf{98.0} & 99.9 & 81.9 & 89.1 \\
Percept-WAM~\cite{han2025percept} & BEV & 98.8 & 98.6 & 94.4 & 99.5 & 84.8 & 90.2 \\
DriveVLA-W0~\cite{li2025drivevla} & -- & 98.7 & \textbf{99.1} & 95.3 & 99.3 & 83.3 & 90.2 \\
AdaThinkDrive~\cite{luo2025adathinkdrive} & Language & 98.4 & 97.8 & 95.2 & \textbf{100.0} & 84.4 & 90.3 \\
\rowcolor{gray!15} READ (VLA setting) & Risk field & \textbf{98.9} & 98.0 & 95.4 & \textbf{100.0} & \textbf{85.5} & \textbf{90.7} \\
\bottomrule
\end{tabularx}
\end{table*}

\section{Experiments}
\label{sec:experiments}

\subsection{Experimental Setup}
\label{sec:exp_setup}

\noindent\textbf{Benchmark and evaluation metrics.}
We evaluate planning on NAVSIM~\cite{dauner2024navsim}, which provides multi-view camera observations, ego state, map information, and high-level navigation commands for diverse urban scenes. Its PDM-Score (PDMS) aggregates no at-fault collision (NC), drivable-area compliance (DAC), ego progress (EP), time-to-collision (TTC), and comfort (C). NAVSIM v2 additionally reports driving-direction compliance (DDC), traffic-light compliance (TLC), lane keeping (LK), human comfort (HC), and ego-state compliance (EC), which are combined into EPDMS. All controlled representation and constraint ablations are conducted on the original NAVSIM protocol so that changes in supervision are compared under the same evaluator.

The aggregate metrics combine hard safety and compliance constraints with continuous driving-quality terms. For NAVSIM v1, all subscores are normalized to $[0,1]$ and
\begin{equation}
\mathrm{PDMS}
= \mathrm{NC}\cdot\mathrm{DAC}\cdot
\frac{5\,\mathrm{EP}+5\,\mathrm{TTC}+2\,\mathrm{C}}{12}.
\label{eq:pdms}
\end{equation}
Thus, collision and drivable-area violations act as multiplicative penalties, whereas progress, time-to-collision, and comfort determine the weighted quality of an admissible trajectory. NAVSIM v2 extends this composition as
\begin{equation}
\begin{aligned}
\mathrm{EPDMS}
&= \mathrm{NC}\cdot\mathrm{DAC}\cdot\mathrm{DDC}\cdot\mathrm{TLC}\\
&\quad\times
\frac{5\,\mathrm{EP}+5\,\mathrm{TTC}
+2\,\mathrm{LK}+2\,\mathrm{HC}+2\,\mathrm{EC}}{16},
\end{aligned}
\label{eq:epdms}
\end{equation}

\noindent\textbf{Models and training.}
For end-to-end (E2E) planning, we use either a ResNet TransFuser-style encoder~\cite{TransFuser} or a ViT initialized from Drive-JEPA~\cite{wang2026drive}. ViT patch tokens are cross-attended by BEV queries, and the shared BEV state supports trajectory and field decoding. E2E models are trained for 40 epochs on two A100 GPUs. For the VLA study, we fine-tune Qwen3-VL-2B~\cite{qwen3technicalreport} with a front-view image, navigation text, risk latents, and trajectory queries for 20 epochs on eight A100 GPUs.

\noindent\textbf{Field settings.}
The field covers $x\in[0,80]$ m and $y\in[-32,32]$ m at 1 m resolution. We predict eight slices at 0.5 s intervals over a 4 s horizon. Map cells, projected future agent boxes, expert waypoints, and kinematic queries are aligned to these slices before their corresponding losses are evaluated. The field decoder and trajectory decoder share the same scene representation and are optimized jointly. Unless stated otherwise, all variants use the same field resolution, ViT backbone, planner, and training schedule.

\subsection{Main Results}
\label{sec:main_results}

\noindent\textbf{Matched end-to-end comparisons.}
Table~\ref{tab:navsim_main} evaluates READ with two end-to-end backbones. Adding READ to TransFuser improves PDMS from 84.0 to 84.6, with gains in NC, DAC, TTC, and EP. With the stronger Drive-JEPA initialization, READ improves PDMS from 89.1 to 90.0 and again raises every non-saturated component. The gain therefore persists from convolutional sensor fusion to a pretrained ViT planner, showing that the field is not tied to one encoder design. In both comparisons, the trajectory decoder and evaluation protocol remain unchanged. The consistent direction of the component scores is therefore more informative than a gain on only one aggregate metric and is compatible with the field acting through the shared representation.

The two matched pairs also separate the proposed supervision from model scale. READ does not replace the encoder, increase the input modalities, or change how trajectories are scored. It changes what the shared spatial state is asked to explain during training. This is the intended role of the field as an intermediate representation: scene features must support both expert motion and a continuous account of nearby risk, while inference remains a direct trajectory prediction problem.

\noindent\textbf{Integration with a VLA planner.}
The VLA experiment tests whether the same field remains useful when planning is mediated by multimodal tokens rather than a BEV-only trajectory decoder. READ reaches 90.7 PDMS and the highest EP among the listed VLA methods. Percept-WAM is likewise query-based but builds on BEV queries, whereas READ structures planning-relevant information through risk-field supervision. READ achieves a higher PDMS in the reported setting, suggesting that field learning provides a complementary way to enrich query-based driving representations. Rather than replacing language-side reasoning, the risk latents complement it with an explicit spatial account of trajectory-relevant geometry and interactions.

\FloatBarrier
\noindent\textbf{Evaluation on NAVSIM v2.} Table~\ref{tab:navsim_v2} reports the complete metric decomposition under the updated protocol. READ reaches 90.1 EPDMS. Under the matched ViT setting, replacing BEV supervision with field learning raises EPDMS by 0.7 and TTC by 0.4, while also improving NC, DAC, DDC, and LK. EP, TLC, and EC remain close. Because the final two rows share the same ViT planner and differ only in intermediate supervision, the comparison isolates the representational effect of field learning. The gains appear across collision-, compliance-, and motion-related components, consistent with field learning improving the shared planning representation under the matched setting. The controlled analyses below use the original NAVSIM protocol.

\begin{table*}[!t]
\centering
\captionsetup{width=0.95\textwidth}
\caption{Planning results on NAVSIM v2. The rightmost EPDMS is the primary aggregate metric under this protocol. The representation column summarizes each method's explicit planning-oriented intermediate representation. The last two rows use the same planner architecture and differ in intermediate supervision.}
\label{tab:navsim_v2}
\scriptsize
\renewcommand{\arraystretch}{1.12}
\setlength{\tabcolsep}{2.0pt}
\begin{tabularx}{0.98\textwidth}{@{}>{\raggedright\arraybackslash}p{1.95cm}>{\centering\arraybackslash}p{1.45cm}*{9}{>{\centering\arraybackslash}X}|>{\centering\arraybackslash}X@{}}
\toprule
Method & Rep. & NC$\uparrow$ & DAC$\uparrow$ & DDC$\uparrow$ & TLC$\uparrow$ & EP$\uparrow$ & TTC$\uparrow$ & LK$\uparrow$ & HC$\uparrow$ & EC$\uparrow$ & \textbf{EPDMS}$\uparrow$ \\
\midrule
Ego Status MLP & Ego state & 93.1 & 77.9 & 92.7 & 99.6 & 86.0 & 91.5 & 89.4 & 98.3 & 85.4 & 64.0 \\
TransFuser~\cite{TransFuser} & BEV & 96.9 & 89.9 & 97.8 & 99.7 & 87.1 & 95.4 & 92.7 & 98.3 & 87.2 & 84.0 \\
HydraMDP++~\cite{Hydra-MDP} & -- & 97.2 & 97.5 & 99.4 & 99.6 & 83.1 & 96.5 & 94.4 & 98.2 & 70.9 & 81.4 \\
DiffusionDrive~\cite{DiffusionDrive} & BEV & 98.2 & 96.2 & 99.5 & 99.8 & 87.4 & 97.3 & 96.9 & 98.4 & 87.7 & 88.2 \\
WoTE~\cite{Li_2025_ICCV} & BEV & 98.5 & 96.8 & 98.8 & 99.8 & 86.1 & 97.9 & 95.5 & 98.3 & 82.9 & 87.7 \\
DriveVLA-W0~\cite{li2025drivevla} & -- & 98.5 & 99.1 & 98.0 & 99.7 & 86.4 & 98.1 & 93.2 & 97.9 & 58.9 & 86.1 \\
MeanFuser~\cite{Wang_2026_CVPR} & BEV & 98.3 & 97.2 & 99.6 & 99.8 & 87.6 & 97.4 & 97.3 & 98.3 & 88.2 & 89.5 \\
\midrule
\makecell[l]{Ours (ViT +\\BEV supervision)} & BEV & 98.7 & 96.8 & 99.5 & \textbf{99.9} & \textbf{87.6} & 98.1 & 97.7 & \textbf{98.4} & 87.3 & 89.4 \\
\rowcolor{gray!15} \makecell[l]{Ours (ViT +\\field learning)} & Risk field & \textbf{99.0} & \textbf{97.2} & \textbf{99.6} & \textbf{99.9} & 87.4 & \textbf{98.5} & \textbf{97.8} & \textbf{98.4} & 87.0 & \textbf{90.1} \\
\bottomrule
\end{tabularx}
\end{table*}

\begin{figure}[!t]
\centering
\includegraphics[width=0.84\columnwidth]{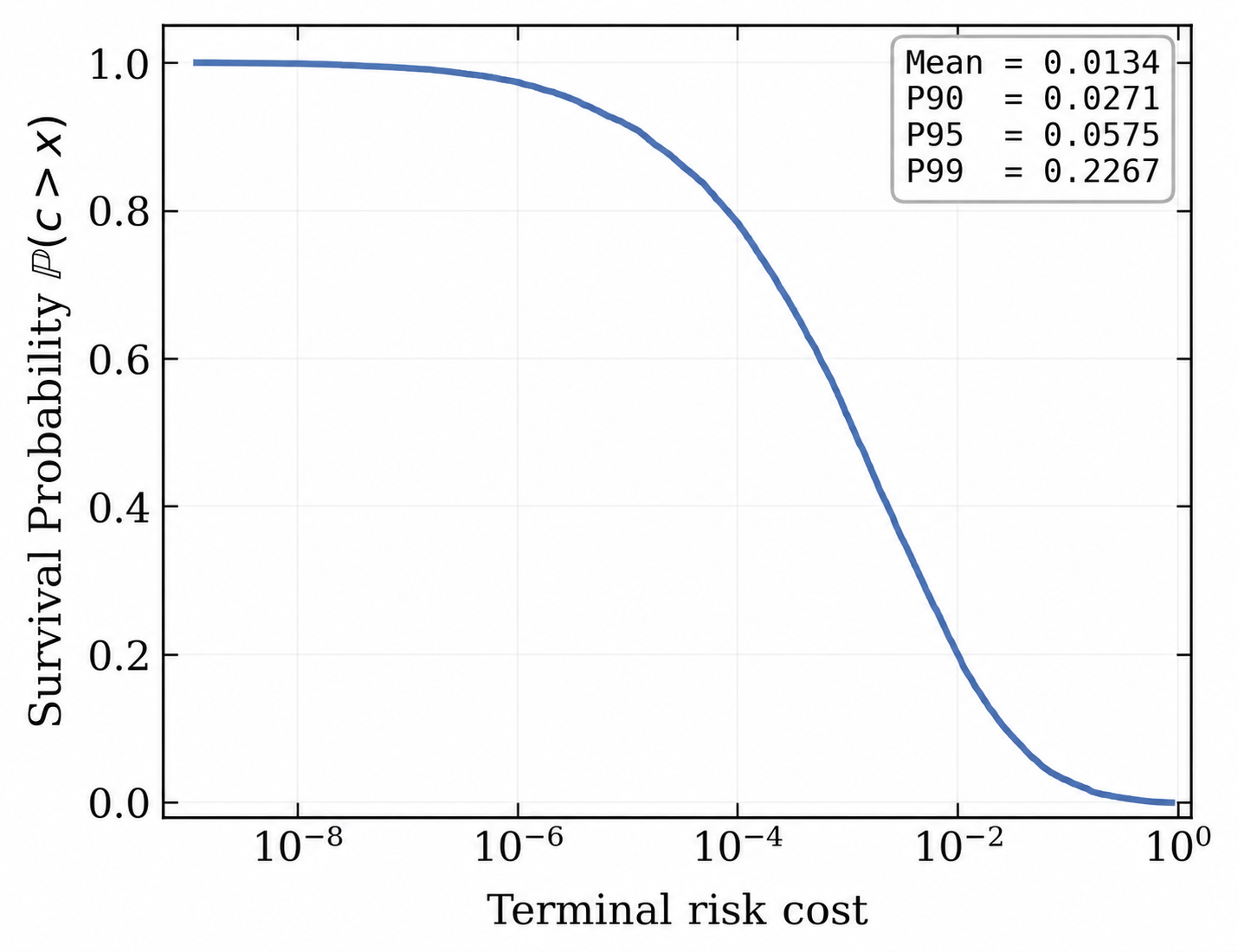}
\caption{Survival function of terminal predicted-trajectory risk on NavTest. Most trajectories predicted by READ terminate in low-risk regions of the learned field.}
\label{fig:consistency}
\vspace{-2mm}
\end{figure}

\begin{figure*}[!t]
\centering
\includegraphics[width=0.94\textwidth]{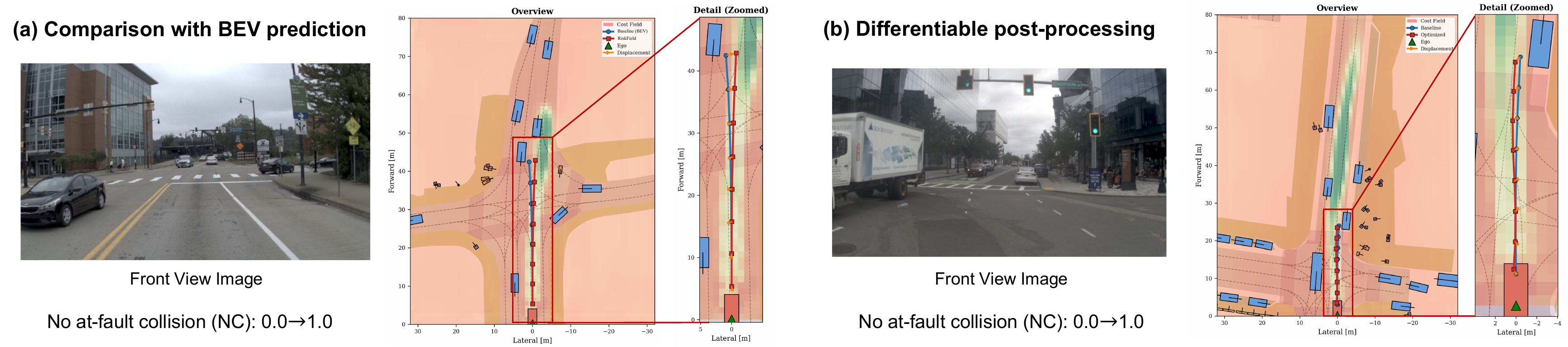}
\caption{Qualitative analysis. (a) READ replaces dense semantic reconstruction with a continuous risk representation under the same planner. (b) In the separate refinement experiment, gradients of the learned field move a selected high-risk prediction while trajectory regularization limits deviation.}
\label{fig:visualization}
\end{figure*}

\subsection{More Analysis and Ablations}
\label{sec:ablation}

\noindent\textbf{Ablation on module combinations.} Table~\ref{tab:ablation_module_combination} compares spatial supervision targets while keeping the ViT backbone, planner, and training schedule fixed. The comparisons ask three increasingly specific questions: whether additional prediction targets help beyond BEV semantics, whether a fixed field target is useful, and whether learning scene-dependent field geometry from partial constraints provides further benefit. BEV semantics is the geometric baseline. Adding a READ decoder to it changes PDMS only slightly (89.06 to 89.23), showing that the gain is not explained by an additional decoder. Eight-step BEV prediction likewise introduces a temporal target with the same horizon as READ but reaches 89.13. This rules out the possibility that the gain of READ merely stems from a longer temporal horizon, since both targets are evaluated over the exactly identical 4‑second planning time window.

\begin{table}[!t]
\centering
\caption{Representation controls under the same ViT planner.}
\label{tab:ablation_module_combination}
\scriptsize
\setlength{\tabcolsep}{2.5pt}
\resizebox{0.92\columnwidth}{!}{%
\begin{tabular}{lccccc}
\toprule
Target & PDMS & NC & DAC & EP & TTC \\
\midrule
BEV semantics & 89.06 & 98.74 & 96.67 & 83.19 & 95.42 \\
BEV + READ field & 89.23 & 98.71 & 96.83 & 83.33 & 95.43 \\
8-step BEV prediction & 89.13 & 98.74 & 96.69 & 83.32 & 95.35 \\
Rule-based field & 89.49 & 98.91 & 96.91 & 83.56 & 95.80 \\
READ field & \textbf{90.02} & \textbf{98.95} & \textbf{97.21} & \textbf{83.75} & \textbf{96.36} \\
\bottomrule
\end{tabular}%
}
\end{table}

The rule-based field uses the analytic field target of FlowDrive~\cite{jiang2025flowdrive}, while retaining our ViT backbone, planner, and training setting. It improves the BEV baseline from 89.06 to 89.49, indicating that explicit field supervision is already useful. In contrast, READ learns the field from partial geometric and behavioral constraints rather than matching a complete, fixed surface. Under the same architecture, READ reaches 90.02 and improves PDMS by a further 0.53, together with NC, DAC, EP, and TTC. The comparison indicates that adapting field geometry to the scene is more effective here than using one rule-defined target across all scenes.

\noindent\textbf{Ablation on risk field supervision.} Table~\ref{tab:ablation_supervision} shows that the constraints provide complementary evidence. Map and future-agent terms anchor annotated hazards, the expert path anchors low field values on demonstrated motion, the kinematic residual queries behavior that departs from the demonstration, and sparsity suppresses unsupported activation. None of these signals defines the complete field alone. Together, the constraints specify where risk should increase, where demonstrated motion should remain low risk, and where unsupported activation should be discouraged, without requiring dense labels for every location in free space.

\begin{table}[!t]
\centering
\caption{Constraint ablation on the original NAVSIM protocol. Each row removes one signal from READ.}
\label{tab:ablation_supervision}
\footnotesize
\begin{tabular*}{0.92\columnwidth}{@{\extracolsep{\fill}}lccccc@{}}
\toprule
Configuration & PDMS & NC & DAC & EP & TTC \\
\midrule
w/o $\mathcal{L}_{\mathrm{agent}}$ & 89.71 & 98.85 & \textbf{97.25} & 83.69 & 95.76 \\
w/o $\mathcal{L}_{\mathrm{ndr}}$ & 89.57 & 98.91 & 96.95 & 83.41 & 96.06 \\
w/o $\mathcal{L}_{\mathrm{reg}}$ & 89.80 & \textbf{98.95} & 97.06 & 83.70 & 96.17 \\
w/o $\mathcal{L}_{\mathrm{gt}}$ & 89.62 & \textbf{98.95} & 96.97 & 83.59 & 95.92 \\
w/o $\mathcal{L}_{\mathrm{res}}$ & 89.73 & 98.93 & 97.07 & 83.61 & 96.07 \\
Full & \textbf{90.02} & \textbf{98.95} & 97.21 & \textbf{83.75} & \textbf{96.36} \\
\bottomrule
\end{tabular*}
\vspace{-2mm}
\end{table}

The strongest and most consistent effect appears in TTC. Removing any constraint reduces TTC from 96.36, with the future-agent term causing the largest drop to 95.76. This pattern directly supports the motivation in Sec.~\ref{sec:intro}: occupancy or free space alone does not determine safety. Two cells can both be physically unoccupied at the current instant while implying different future collision margins because of nearby motion. Future-agent and behavior-derived constraints provide the temporal evidence needed to organize this free space differently. The PDMS changes are smaller because the metric combines several partly saturated components, but every removal lowers the aggregate score. We therefore interpret the table as evidence for complementary constraint roles, with TTC giving the clearest safety-related validation.

The low-cost constraints are equally important to this interpretation. Expert waypoints do not label every other location as unsafe. They only anchor one demonstrated feasible path, while sparsity discourages the field from assigning unsupported risk everywhere else. The unqueried free space is therefore not forced into a binary safe or unsafe class. Its value is determined through the smooth basis and the scene representation, which is precisely the flexibility lost when a complete field is specified before learning.

\noindent\textbf{Effect of regularization weight.} Table~\ref{tab:ablation_reg_weight} shows the effect of the sparsity weight $\lambda_r$. A moderate value of 0.5 performs best: larger weights encourage sparsity but may suppress safety-critical support, whereas smaller values allow risk to spread into unsupported regions.

\noindent\textbf{Consistency between predicted trajectories and risk field.}
To examine alignment between the learned field and planner behavior, we evaluate the terminal risk of every predicted trajectory on NavTest. Figure~\ref{fig:consistency} shows that most predictions terminate in near-zero-risk regions, while only a small long-tailed subset has non-negligible risk. This suggests that the planner generally selects trajectories aligned with the learned risk landscape and that the remaining high-risk cases are candidates for selective refinement. We use this distribution as a consistency diagnostic rather than as calibrated collision probability or a guarantee of safety.

%The distribution has a median of $0.0134$; its 90th and 95th percentiles are $0.0271$ and $0.0575$, respectively. Only the far tail rises substantially, reaching $0.2267$ at the 99th percentile.

\noindent\textbf{Risk field as a differentiable post-processing module.} We further evaluate the learned field as an optional refinement module for trajectories generated by external planners. The planner is frozen and only its output waypoints are optimized. Table~\ref{tab:posthoc_field_opt} shows that READ improves TransFuser from 84.14 to 84.75 PDMS and DiffusionDrive from 88.08 to 88.43, with gains in every displayed component. The gains for both regression-based and diffusion planners indicate that the learned field serves as a planner-agnostic post-hoc guide for candidate trajectories.

\begin{table}[!t]
\centering
\caption{Sensitivity to the sparsity weight $\lambda_r$.}
\label{tab:ablation_reg_weight}
\footnotesize
\begin{tabular*}{0.92\columnwidth}{@{\extracolsep{\fill}}lccccc@{}}
\toprule
$\lambda_r$ & PDMS & NC & DAC & EP & TTC \\
\midrule
1.0 & 89.66 & 98.88 & 97.04 & 83.59 & 96.00 \\
0.5 & \textbf{90.02} & \textbf{98.95} & \textbf{97.21} & \textbf{83.75} & \textbf{96.36} \\
0.1 & 89.29 & 98.86 & 96.72 & 83.22 & 95.92 \\
\bottomrule
\end{tabular*}
\end{table}

\subsection{Qualitative Analysis}

\noindent\textbf{Comparison with BEV semantic supervision.}
Figure~\ref{fig:visualization}(a) compares the two intermediate targets under the same planner. BEV supervision reconstructs semantic categories, while READ organizes the scene as continuous planning-relevant influence around static structure and dynamic agents. The field extends into nearby free space without assigning one fixed analytic profile to every object. This difference is most visible where semantic occupancy alone is ambiguous for motion. READ can represent a graded margin around a source and vary that margin with the surrounding configuration, while retaining a continuous surface that can be queried at arbitrary trajectory points. This qualitative difference mirrors the controlled comparison in Table~\ref{tab:ablation_module_combination}: the benefit of READ is not a denser semantic reconstruction, but a scene-conditioned representation that adapts to the local driving context.

\noindent\textbf{Trajectory refinement visualization.}
Figure~\ref{fig:visualization}(b) visualizes the optional procedure evaluated only in Table~\ref{tab:posthoc_field_opt}. The initial trajectory intersects a high-value region. Successive updates follow local field gradients, while trajectory regularization preserves smoothness and forward progress. The update therefore balances two requirements: reducing exposure to the learned field and preserving the intent of the original prediction. The example illustrates how a learned representation can be queried and optimized directly without changing the frozen planner. It complements the quantitative table but is not used to claim field calibration or closed-loop safety.

\begin{table}[!t]
\centering
\caption{Optional trajectory refinement with frozen planners.}
\label{tab:posthoc_field_opt}
\scriptsize
\setlength{\tabcolsep}{2.6pt}
\resizebox{0.95\columnwidth}{!}{%
\begin{tabular}{lccccc}
\toprule
Configuration & PDMS & NC & DAC & EP & TTC \\
\midrule
TransFuser & 84.14 & 97.77 & 92.78 & 78.92 & 93.30 \\
$\quad$+ READ refinement & \textbf{84.75} & \textbf{97.82} & \textbf{93.31} & \textbf{79.39} & \textbf{93.60} \\
DiffusionDrive & 88.08 & 98.15 & 96.27 & 81.19 & 94.77 \\
$\quad$+ READ refinement & \textbf{88.43} & \textbf{98.16} & \textbf{96.51} & \textbf{82.41} & \textbf{95.10} \\
\bottomrule
\end{tabular}%
}
\end{table}

\section{Conclusion}
\label{sec:conclusion}

We introduced READ, a risk-aware planning framework that learns a differentiable spatiotemporal field from complementary geometric and behavior-derived constraints. The field provides a compact, planning-aligned representation beyond dense BEV reconstruction and can be integrated into both end-to-end and VLA planners. Matched NAVSIM experiments show consistent improvements, while the rule-based control confirms both the value of explicit field supervision and the additional benefit of adaptive field geometry. The learned field can also serve as an optional refinement module for trajectories generated by external planners.

\textbf{Limitations and future work.}
The current dense spatiotemporal field may introduce unnecessary computation in large-scale settings, motivating sparse or adaptive risk representations. Future work could also explore richer supervision signals, such as intervention or takeover data, for modeling safety-critical driving behaviors.

\bibliographystyle{IEEEtran}
\bibliography{references}

\end{document}